# Deep Learning Approach for Classifying the Aggressive Comments on Social Media: Machine Translated Data Vs Real Life Data

Mst Shapna Akter*, Hossain Shahriar†, Nova Ahmed ‡, Alfredo Cuzzocrea§

*Department of Computer Science, Kennesaw State University, USA
† Department of Information Technology, Kennsaw State University, USA
‡ Department of Electrical and Computer Engineering (ECE), North South University, Bangladesh
§ iDEA Lab, University of Calabria, Rende, Italy

{*makter2@students.kennesaw.edu,† hshahria@kennesaw.edu, ‡ nova.ahmed@northsouth.edu, § alfredo.cuzzocrea@unical.it }

*Abstract*—Aggressive comments on social media negatively impact human life. Such offensive contents are responsible for depression and suicidal-related activities. Since online social networking is increasing day by day, the hate content is also increasing. Several investigations have been done on the domain of cyberbullying, cyberaggression, hate speech, etc. The majority of the inquiry has been done in the English language. Some languages (Hindi and Bangla) still lack proper investigations due to the lack of a dataset. This paper particularly worked on the Hindi, Bangla, and English datasets to detect aggressive comments and have shown a novel way of generating machine-translated data to resolve data unavailability issues. A fully machine-translated English dataset has been analyzed with the models such as the Long Short term memory model (LSTM), Bidirectional Long-short term memory model (BiLSTM), LSTM-Autoencoder, word2vec, Bidirectional Encoder Representations from Transformers (BERT), and generative pre-trained transformer (GPT-2) to make an observation on how the models perform on a machine-translated noisy dataset. We have compared the performance of using the noisy data with two more datasets such as raw data, which does not contain any noises, and semi-noisy data, which contains a certain amount of noisy data. We have classified both the raw and semi-noisy data using the aforementioned models. To evaluate the performance of the models, we have used evaluation metrics such as F1-score, accuracy, precision, and recall. We have achieved the highest accuracy on raw data using the gpt2 model, semi-noisy data using the BERT model, and fully machine-translated data using the BERT model. Since many languages do not have proper data availability, our approach will help researchers create machine-translated datasets for several analysis purposes.

## I. INTRODUCTION

The invention of the World Wide Web (WWW) had a significant influence on social media as users may share a variety of content, including instructive, entertaining, and personal information, very fast without being present physically but using only a digital device [1]. Facebook, Twitter, Instagram, and YouTube are the most widely used social media networks [2]. These platforms enable users to share their ideas, knowledge, and point of view. However, those platforms also have a negative side [3]. In some cases, freedom in digital social media results in detrimental effects, when not well-used, including despair, depression, and even suicide [4, 5]. As a result, social media is becoming riskier for users and may even encourage some to end their lives. Therefore, hate speech has been the subject of numerous investigations to determine the causes and actions against online hostility .

Cyberbullying is the behavior of repetitive hurting of an individual or group of individuals by the dissemination of offensive content or the use of other types of social violence through the use of digital media [6–8]. Mostly, teenagers experience cyberbullying on social media [9]. One study has shown that 36.5% of students have dealt with cyberbullying at least once in their lives, and among all the other forms of online comments, rude or cruel remarks were the most prevalent. Another study discovered that out of 1,501 adolescents in the USA between the ages of 10 and 17, 12% admitted to abusing someone online, 4% admitted to being the victims, and 3% admitted to being both the aggressor and the victim of cyberbullying [10]. A survey by Sri Lanka's law enforcement agency, the Cyber Crimes Division (CID), reveals that more than 1000 incidences of cyberbullying were recorded there. Over 90% of university students reported having experienced cyberbullying, and almost all poll respondents stated they knew someone who had been bullied online. 80% of the cyberbullying experienced by Sri Lankans occurred on Facebook. Inconvenient videos or photographs have been posted online by 65% of college students. 15% of users posted personal information online, 9% disseminated inaccurate information about others and lies, and 2% posted offensive material [11]. A broad audience and extended visibility period come with transmitting cyberbullying faster and easily, which is a big problem nowadays [12]. It becomes an everyday occurrence, and victims face it repetitively, which causes both mental and physical issues [13]. Schneider et al. [14] showed a relationship between victimization and five categories of physical distress in a survey of MetroWest Adolescent Health by collecting information from over 20,000 pupils. Among the cyberbullying victims, self-harm (24%) and depressed symptoms (34%) were the highest rates of psychological distress.

Since the number of users grows daily, a thorough investigation is now required to address the problem of cyberbullying. Previous investigations in this field lack many criteria, such as accurate detection with effective algorithms and data unavailability for training the advanced AI technology, which is crucial to address as soon as possible [15, 16]. Previously, Perera et al. [11] conducted an investigation to detect cyberbullying on social media. However, they used low instances of a dataset ( 1000 labeled text data) for making classification with classifiers such as Support Vector Machine ( SVM), which resulted in a very low accuracy ( 74%). Later, Alotaibi et al. [17] proposed a multichannel deep learning framework for cyberbullying detection on social media using a 55,788 Twitter dataset. They developed Multichannel deep learning, which did not provide satisfactory results ( 88% accuracy). Finally, Ahmed et al. [18] used Deep Neural Network to detect cyberbullying on social media using 44001 comments From Facebook. They also tried to develop a Hybrid Neural Network, but the result was ineffective ( 85% accuracy).

However, countries like Bangladesh and India lack proper investigations due to the lack of data availability. None of the investigations has been done to resolve the data unavailability issues. Therefore, those countries are more vulnerable to cyberbullying and cyber aggression due to the lack of research and inappropriate action. We have focused on Hindi, Bangla, and English to detect aggressive comments on the social media platform. We have used the TRAC-2 dataset, which contains English, Bangla, and Hindi comments.

Furthermore, we have tried to resolve the data unavailability issues by creating a fully machine-translated English dataset. Since the dataset is very important to learn a model for detecting aggressive comments, if one language lacks the dataset, it may become impossible to solve the aggressive comment issue for that language. Google Translator can translate data from one language to another but contains a lot of noises that may not be appropriate for training a deep learning model. In this paper, we have shown how the deep learning models perform on machine-translated noisy datasets and compared the result with the raw and semi-noisy datasets, which have provided a clear observation of how the deep learning models perform on noise-free, semi-noisy, and fully-noisy datasets. The raw dataset contains an imbalanced data issue. We have used a machine translated augmentation process to avoid the imbalanced data problem and constructed a semi-noisy dataset. After constructing the semi-noisy and fully-noisy datasets, we extracted features using the Bert embedding model. The extracted features have been fed to the deep learning models such as LSTM, BiLSTM, LSTM-Autoencoder, Word2vec, BERT, and gpt2 model for classifying the aggressive comments. We have checked the performance of the models using performance metrics such as f1-score, precision, recall, and accuracy.

All of the evaluations have been done on the unseen dataset. We have got the highest result accuracy of 80 percent on English raw data and 73 percent on Bangla raw data using the gpt2 model. For the semi-noisy dataset, we have got highest accuracy of 75 percent on English, 71 percent on Bangla, and 68 percent on the Hindi dataset using the BERT model. For the fully machine-translated English dataset, we have achieved the highest accuracy of 78 percent using the BERT model. The main contributions of this paper can be summarized in three aspects as follows.

(1) A process of generating synthetic data has been shown. We have also shown the performance of traditional models such as LSTM, BiLSTM, LSTM-Autoencoder, Word2vec, BERT, and GPT2 models on synthetic data. We named the synthetic data "noisy" since the data is machine translated and contains many noises.

(2) Extensive experiments on three kinds of datasets: noisy, semi-noisy, and noise-free using traditional models such as LSTM, BiLSTM, LSTM-Autoencoder, and Word2vec, BERT, and GPT2 has been performed to make a comparison using the evaluation metrics such as F1-score, precision, recall, and accuracy. The semi-noisy data refers to the combination of noisy data and raw data. This process is essential to compare different neural network models using different intensities of noise levels present in the dataset.
Following our approach, suicidal activities may reduce to an extent by detecting the aggressive comments and taking action based on the prediction. Our approach can be useful for languages that lack data availability.

The rest of this paper is arranged as follows. Section 2 provides the background needed for the study. The data sources, preprocessing methods, and models used in this work for aggression detection tasks are discussed in Section 3. The simulation results based on the classification algorithms and the comparison using the derived results are analyzed in Section 4. Finally, this paper is summarized in Section 5.

## II. RELATED WORK

Deep learning models such as Word2vec, LSTM, BiLSTM, BERT, XLM-Roberta, and FastText are popular models for dealing with textual data. Some models can capture the true meaning of the sentence very well; some require a lesser computational cost. Many of these models have been used for cyberbullying detection. Therefore, we have tried to go through very recent as well as primitive investigations that happened, particularly in cyberbullying field.

Previously Perera et al. [11] showed an approach for accurate cyberbullying detection and prevention on social media using 1000 manually labeled texts from Twitter. They labeled the dataset manually as some comments may contain slang words but still can be non-bullying comments, e.g. "you have done fucking well in the exam". They tried to understand the true meaning and annotated it accordingly. They used Support Vector Machines (SVM) for classification and Logistic regression to select the best combination of features. Their proposed solution provides 74% accuracy.

Simon et al. [19] showed a systematic review of machine learning trends in the automatic detection of hate speech on social media platforms. A total of 31,714 articles from 2015 to 2020 were examined; 41 papers were included based on inclusion criteria, while 31,673 papers were excluded according to exclusion criteria. This study concluded that machine learning and deep learning are the most successful methods for classifying hate speech on social media. Moreover, they found that many researchers used the support vector machine algorithm for classification, while the deep learning models are also gaining popularity daily. A similar systematic review was shown by Castaño-Pulgarín et al. [20]. They found 67 studies eligible for analysis out of 2389 papers in the online search. They showed a qualitative study but did not make any analysis of technical approaches. The results showed that the victims are mainly from Muslim countries, and the abuser targets the Muslim religion for making hate speech.

Roy et al. [21] used Multilingual Transformers for hate speech detection. They examined the issue of identifying offensive and hateful words on Twitter. They specifically tried to address two classification issues. First, categorize each tweet as either hostile and insulting (class HOF) or not (class NOT). Second, categorize it into one of the following three categories: hate speech, offensive, and profanities (HATE, OFFN, PRFN). They used the XLM-Roberta classification model. They achieved Macro F1 scores of 90.29, 81.87, and 75.40 for English, German, and Hindi, respectively, while performing hate speech detection and 60.70, 53.28, and 49.74 during fine-grained classification.

Alotaibi et al. [17] proposed a multichannel deep learning framework for detecting cyberbullying on social media. This method divides Twitter comments into categories such as aggressive and non-aggressive categories. They classified the comments using algorithms such as transformer, bidirectional gated recurrent, and convolutional neural network. The effectiveness of the suggested strategy was evaluated using a combination of three well-known hate speech datasets. The proposed approach had an accuracy rate of about 88%.

Sadiq et al. [22] used a deep neural model to detect aggressive comments on Twitter. They used a Multilayer Perceptron and fed manually engineered features onto it. They also experimented with a cutting-edge CNN-LSTM and CNN-BiLSTM deep neural network combination; both of them worked well. Their statistical findings demonstrated that the proposed model worked best with 92% accuracy detecting aggressive behavior.

Ahmed et al. [18] used Deep Neural Network to detect cyberbullying on social media. The dataset they used comprises of 44001 user comments from Facebook sites. They categorized the datasets into categories such as religious, troll, threat, non-bully, and sexual and preprocessed the information to remove errors like incorrect punctuation and flawed characters before feeding it into the neural network. The pre-processing procedures were carried out in three stages: removing stop words, tokenizing string, and converting padded sequence. Their model consists of three parts: 1. identifying harassment related comments which contains descriptors such as threat, troll, and religious as bully, 2. using hybdrid classification model for categoeizing all five classes, 3. using an ensemble approach for increasing accuracy by pooling the predicted results from the multiclass classification models. The model provides 85% accuracy, while their binary classification model provided 87.91% accuracy.

Kumar and Sachdeva [23] showed a Bi-GRU with attention and CapsNet hybrid model for cyberbullying detection on social media. They demonstrated their proposed model's result and showed that for MySpace and Formspring, the F-score has improved by almost 9% and 3%, respectively.

Alam et al. [24] showed an ensemble-based machine learning approach for detecting cyberbullying. They developed both single and double-voting models to classify offensive and non-offensive comments. The dataset collected from Twitter. To compare their result they used four machine learning models, three ensemble models, and two feature extraction methods. Moreover, they coupled various n-gram analyzers with those models. The result showed that their proposed SLE and DLE voting classifiers performed best among all the aforementioned models. The most outstanding performance for their suggested SLE and DLE models was 96% when TFIDF (Unigram) feature extraction was used with K-Fold cross-validation.

Desai et al. [25] used a machine-learning approach to detecting cyberbullying on social media. They proposed a model based on certain characteristics that should be considered when identifying cyberbullying and applied a few characteristics with the aid of a bidirectional deep learning model known as BERT. They split the features into sentimental, syntactic, sarcastic, semantic, and social. Their suggested model performed more accurately (91.90% accuracy), which was a better result when compared to the typical machine learning models employed on comparable datasets.

III. METHODOLOGY

A. Dataset Specification

The dataset used in this work is collected from Trac-2 (workshop on trolling, Aggression, and cyberbullying), which contains 25,000 comments from three social media – Facebook, Youtube, and Twitter, in three languages– English, Bengali, and Hindi [26]. The shared task has two groups: Sub-Task A (Aggressive comments) and Sub-Task B ( misogyny comments). Sub-Task A comprises three classes: Non-Aggressive (NAG), Overtly Aggressive (OAG), and Covertly-Aggressive (CAG). The indirect aggressive comments are annotated as Covertly-Aggressive (CAG), the direct aggressive comments are annotated as Overtly-Aggressive (OAG), and no aggressive comments are annotated as Non-aggressive (NAG). Likewise, Sub-task B contains two classes: GEN and NGEN. A comment which indicates a man, woman, or transgender is annotated as GEN, and a comment that does not indicate gender is annotated as NGEN. All three Datasets contain both train and test sets. In our project, we experimented with Sub-Task A, since Sub-task A's features fully align with our purpose of prediction, which is cyberbullying detection. The statistics of the dataset provided by the organizations have been shown in Table 1 for Sub-Task A.

Table 1 Label distribution of dataset for Sub-Task A

| Authors | Data Source | Instance | Methods | Limitation |
|---|---|---|---|---|
| Set | NAG | OAG | CAG | Total |
| English Training | 3375 | 453 | 435 | 4263 |
| English Testing | 836 | 117 | 113 | 1066 |
| Hindi Training | 2245 | 829 | 910 | 3984 |
| Hindi Testing | 578 | 211 | 208 | 997 |
| Bangla Training | 2078 | 898 | 850 | 3826 |
| Bangla Testing | 522 | 218 | 217 | 957 |

Some examples of the text data is shown in Figure 1.

*B. Data Preprocessing*

To create semi-noisy data, we have added noises with the raw data until the dataset resolves imbalanced issues. The data we have used is initially highly imbalanced for sub-task A. The imbalanced data perform poorly for predicting the aggressive comments. The category NAG holds 50 percent of the total text data, and the category OAG and CAG hold the other 50 percent of total text data. To resolve the imbalanced data issue, we have augmented the text data in such a way that all of the classes maintain almost the same amount of text data. We have adopted two methods for the augmentation process– Noise Addition and Data Translation.

Noise Addition: Noises are added by replacing a word with synonyms or antonyms, adding random stop words, and shuffling some words on raw text data. The process helps to increase the corpus size while keeping the context the same as the raw dataset.

Data Translation: Data have been translated from one language to another, e.g. English to Bangla, using google translator. All the languages have the same sub-tasks and classes. So, we have translated the texts with all possible combinations until the dataset reaches a balanced position. We translated the texts for NAG, OAG, and CAG classes from Sub-Task A.

Fig. 1. Example of dataset with categories

We have added texts from the noise augmentation process and texts from the translation augmentation process into the raw data so that the new corpus holds an almost equal number of text data for Sub-task A. The statistics of the dataset after adding the augmented data with raw data shown in Table-2 for Sub-Task A.

TABLE 2 Label distribution for Augmented + Raw dataset used for Subtask: A

| Set | NAG | OAG | CAG | Total |
|---|---|---|---|---|
| English Training | 3375 | 2251 | 2546 | 8172 |
| English Testing | 836 | 117 | 113 | 1066 |
| Hindi Training | 2245 | 3497 | 1810 | 7552 |
| Hindi Testing | 578 | 211 | 208 | 997 |
| Bangla Training | 2078 | 1959 | 1966 | 6003 |
| Bangla Testing | 522 | 218 | 217 | 957 |

*C. Fully Machine Translated Data*

Using the translation augmentation process, we have generated a complete machine-translated English dataset which is fully noisy. We have translated the Bangla and Hindi Sub-Task A texts into English for creating the new dataset. The statistics of the fully translated data is shown in Table 3 for Sub-Task A.

TABLE 3 : Label distribution for fully translated English dataset for Subtask:A

| Set | NAG | OAG | CAG | Total |
|---|---|---|---|---|
| English Training | 4373 | 3096 | 2588 | 10057 |

Some examples of the Machine translated English data is shown in Figure 2.

| | |
|---|---|
| It doesn't matter if Moumita Sarkar or Hindu Der Modda is extra-married Kora.Well said, brother | NAG |
| What are you looking at? 200 rupees has been said to be 1000 rupees, there is a mobile phone ,সব nGreen has come face to face. You read everything from your reading, কিছু ndont want to do anything? Here is what we have said beforeDarun Diacho Dada Sothi Ranu Di is doing very wrong . | NAG |
| Kunfu comming from china and kutta comming from Pakistan.. | OAG |
| Disliked n unsubscribed. Ohh wait, I wasn't subscribed. | OAG |
| I hat ranu Mondal | CAG |
| One word for all haters go die somewhere else.. | CAG |

Fig. 2. Example of fully translated English dataset with categories

### D. Input Representation

The raw text, semi-noisy, and fully noisy datasets are converted into a machine understandable number representation. The computer can only understand the numbers; so, it is necessary to convert the text into numbers before feeding it into the models. We have used a BERT tokenizer for BERT models and TensorFlow.Keras tokenizer for Autoencoder, LSTM, and BiLSTM models.

### E. Classification Models

The LSTM, BiLSTM, LSTM-autoencoder, Word2vec, BERT models, and GPT-2 have been applied to the text data of the Trac-2 workshop. All of the models are individually applied on English, Bangla, and Hindi datasets. For training the models, the dataset has been split into two parts: training and validation. Finally, all of the trained models are evaluated on the test dataset.

*1) LSTM:* LSTM is A popular artificial Recurrent Neural Network (RNN) model, which works with datasets that preserve sequences such as text, time series, video, and speech. Using this model for sequential datasets is effectiveSince as it can handle single data points. It follows the Simple RNN model's design and an extended version of that model. However, unlike Simple RNN, it has the capability of memorizing prior value points since it developed for recognizing long-term dependencies in the text dataset. Three layers make up an RNN architecture: input layer, hidden layer, and output layer [27, 28]. Figure 3 depicts the LSTM's structural layout. The elementary state of RNN architecture is shown as the mathematical function:

$$h_t = f(h_{t-1}, x_t; \theta) \quad (1)$$

Here, $\theta$ denotes the function parameter, $h_t$ denotes the current hidden state, $x_t$ denotes the current input, $f$ denotes the function of previous hidden state.

When compared to a very large dataset, the RNN architecture's weakness is the tendency to forget data items that are either necessary or unneeded. Due to the nature of time-series data, there is a long-term dependency between the current data and the preceding data. The LSTM model has been specifically developed to address this kind of difficulty. It is first proposed by Hochreiter Long [29].

This model's main contribution is its ability to retain long-term dependency data by erasing redundant data and remembering crucial data at each update step of gradient descent [30]. The LSTM architecture contains four parts: a cell, an input gate, an output gate, and a forget cell [31].

The purpose of the forget cell is to eliminate extraneous data by determining which data should be eliminated based on the state (t) − 1 and input x(t) at the state c(t) − 1.

At each cell state, the sigmoid function of the forget gate retains all 1s that are deemed necessary values and eliminates all 0s that are deemed superfluous.[29, 32, 33]. The forget gate state's equation is stated as follows:

$$f_t = \sigma(W_f \cdot [h_{t-1}, x_t] + b_f) \quad (2)$$

where $f_t$ denotes sigmoid activation function, $h_{t-1}$ denotes output from previous hidden state, $W_f$ denotes weights of forget gate, $b_f$ denotes bias of forgetting gate function, and finally $x_t$ dentoes current input.

After erasing the unneeded value, new values are updated in the cell state. Three steps make up the procedure: The first step is deciding which values need to update using sigmoid layer called the "input gate layer". Second, creating a vector of new candidate values using the tanh layer. Finally, steps 1 and 2 are combined together to create and update the state.

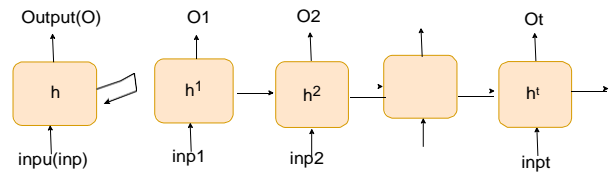

Fig. 3. LSTM neural network structure.

The equation for the sigmoid layer is as follows:

$$i_t = \sigma(W_i \cdot [h_{t-1}, x_t] + b_i) \quad (3)$$

producing a new value to the state $C(t)$, while the sigmoid determines which value should be updated.
The equation for tanh layer's equation is as follows:

$$\tilde{C}(t) = \tanh(W_c \cdot [h_{t-1}, x_t] + b_C) \quad (4)$$

The addition of $C(t) * i_t$ and $C_{t-1} * f_t$ updates the new cell at state $C(t)$. The updated state's equation is:

$$C_t = C_{t-1} * f_t + \tilde{C}(t) * i_t \quad (5)$$

In order to determine which output needs to be maintained, the output is ultimately filtered out using the sigmoid and the tanh functions.

$$O_t = \sigma(W_o \cdot [h_{t-1}, x_t] + b_o) \quad (6)$$

$$h_t = O_t * \tanh(C_t) \quad (7)$$

In this state, $h_t$ gives outputs that are used for the input of the next hidden layer.

*2) BiLSTM:* The Bidirectional Long short-term memory (BiLSTM) was first proposed by GRAVES [34]. The architecture of BiLSTM can learn patterns from both past-to-future and future-to-past data. This idea sets it apart from the LSTM model, which can learn patterns from the past to the future. Figure 4 depicts the bidirectional LSTM's structural layout. The backward propagation layer primarily functions as a forwarding LSTM reverse layer. The hidden layer synthesizes information from both the forward and backward directions [35]. As a result, "the reverse direction of forwarding direction" is used to calculate the LSTM reverse layer. The formula for computing the BiLSTM network is:

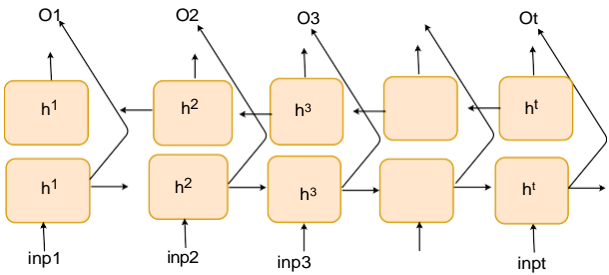

Fig. 4. BiLSTM neural network structure.

The mathematical equation [35] of backward propagation is as follows:

$$h_f = f(w_{f1} x_t + w_{f2} h_{t-1}) \quad (8)$$

$$h_b = f(w_{b1} x_t + w_{b2} h_{t+1}) \quad (9)$$

Where, $h_f$ denotes as the forward layer's output, and $h_b$ denotes as the reverse layer output.
The output derived from the hidden layer is given below:

$$O_i = g(w_{o1} * h_f + w_{o2} * h_b) \quad (10)$$

*3) LSTM-Autoencoder:* An LSTM Autoencoder is an autoencoder implemented for sequential data by following an Encoder-Decoder LSTM architecture. For a given sequential dataset, LSTM-Autoencoder is designed to read the input sequences, encode the sequence, decode the sequence, and reconstruct the sequence. The model's performance is estimated based on the ability to how correctly it can reconstruct the sequence. LSTM autoencoder can be used on video, text, audio, and time-series sequence data [36].

*4) Word2vec:* Word2vec is a word embedding model which deals with textual data. The representation of a word is very complex and can not be understood by machine learning algorithms. Therefore, word embedding makes it easy to align the presentation of the words in such a way that it preserves each word's meaning. The model maps each word into vectors of real numbers using a neural network model and is capable of capturing a long sequence of semantic and syntactic relationships. The word2vec is built with a two-layered neural network. It can detect synonymous words and suggest additional words for partial sentences [37]. Based on a corpus of text, they are used to build and train semantic vector spaces, which frequently have several hundred dimensions. Each word from the corpus is represented as a vector in this space. In this area, words that have similar contexts are situated next to one another. Two methods can serve as the foundation for the word2vec architecture: a continuous bag-of-words model or a continuous skip-gram model (CBOW). While the latter uses the current word to predict surrounding words, the former uses the context to predict the current word. Both models have a low computational complexity, making it possible for them to quickly process a corpus with a size in the billions of words. Although CBOW models are quicker, skip-gram has been found to perform better on short datasets. Therefore, we decided to use the latter model.

*5) BERT:* Bidirectional Encoder Representations from Transformers (BERT) is a pre-trained model which has been trained on a sizable unsupervised Wikipedia or Book corpus. It ensures a deeper sense of language context as it learns text sequence from left to right or a combination of left to right and right to left. Therefore, it doesn't stick with one single direction [38]. A pre-trained BERT model can be improved with just one more layer to provide cutting-edge outcomes in a variety of NLP tasks. For BERT models, there are two variations: BERTBASE and BERTLARGE. This paper has used two types of BERT models: BERT base and BERT MultiLingual. Since the BERT model is trained on English text data, leaving low-resource languages such as Bangla and Hindi behind. Whereas, the Multilingual Bert model was trained on Wikipedia content with a shared vocabulary across all languages, which supports 104 languages. Bangla and Hindi dataset has been classified using the Multilingual Bert model [39, 40].

Input Format: A certain format for the input token sequence is required by BERT. Every sequence should begin with a [CLS] (classification token), and each sentence should be followed by a [SEP] (separation token). The sequence embedding that can be used to classify the entire sequence is the output embedding that corresponds to the [CLS] token.

*6) GPT-2:* Generative pretrained transformer-2 (GPT-2) is one of the most standard states of the art generative modeling transformers. It has been trained on a large web text corpus. GPT2 is mostly used for the next sequence prediction, question answering, sequence classification, abstract, or text summarization. GPT-2 is known as a transformer decoder as it does not construct with lots of encoders; instead, it relies mainly on decoders as its main structural framework. GPT2 has many variances, such as GPT-2 SMALL, GPT-2 MEDIUM, GPT-2 LARGE, and GPT-2 EXTRA LARGE. We have used GPT-2 MEDIUM for classifying the aggression detection [41].

*F. Evaluation metrics*

Evaluating a model's performance is necessary since it shows how close the model's predicted outputs are to the corresponding expected outputs. The evaluation metrics are used to evaluate a model's performance. However, the evaluation metrics differ with the types of models. The types of models are classification and regression. Regression refers to the problem that involves predicting a numeric value. Classification refers to the problem that involves predicting a discrete value. The regression problem uses the error metric for evaluating the models. Unlike the regression problem model, The classification problem uses the accuracy metric for evaluation. Since Our motive is to classify the aggressive comments, we used Accuracy and F1 score as the main Evaluation metric[42].

Precision : When the model predicts a positive result, it should specify how much the positive values are correct. Precision is used when the false positives are high. In aggressive detection classification, if the model gives low precision, many comments are said to be aggressive; for high precision, it will ignore the False positive values by learning with false alarms. The precision can be calculated as follows:

$$Precision = \frac{TP}{TP + FP} \quad (11)$$

Recall : Recall is the opposite of Precision. Precision is used when the false Negatives are high. In the aggressive detection classification problem, if the model gives low recall, many comments are said as non-aggressive; for high recall, it ignores the FalseNegative values by learning with false alarms. The recall can be calculated as follows:

$$Recall = \frac{TP}{TP + FN} \quad (12)$$

F1 score: F1 score combines precision and recall and provides an overall measure of the models' accuracy. The value of the F1 score lies between 1 and 0. If the predicted value matches with the expected value, then the f1 score gives 1, and if none of the values matches with the expected value, it gives 0. The F1 score can be calculated as follows:

$$F1 = \frac{2 \cdot precision \cdot recall}{precision + recall} \quad (13)$$

Accuracy : Accuracy determines how close the predicted output is to the actual value.

$$Accuracy = \frac{TP + TN}{TP + TN + FP + FN} \quad (14)$$

here, TN refers to True Negative and FN refers to False Negative.

IV. RESULT AND DISCUSSION

Sub-Task A has been considered for classifying the aggression comments on social media. Since the dataset is imbalanced, several data preprocessing methods is adopted before classifying the aggression detection. Data augmentation, such as machine translation and noise addition, are used to resolve the imbalanced data issues. Finally, a fully machine-translated English data has been created to check the performance of existing deep learning models on machine-translated data that contains most noises. We have trained five deep learning models: LSTM, BiLSTM, LSTM-autoencoder, Word2vec, BERT transformer, and GPT-2 models using all datasets. The models are evaluated using evaluation metrics such as F1 score, accuracy, precision, and recall. We derived the metric evaluation result over the unseen test dataset. The performance of the models is shown in Table 4 and Table 5, and Table 6. The classification results for Autoencoder, LSTM, BiLSTM, word2vec, and Bert models using raw data, raw data with augmented data, and machine-translated English data are shown below:

TABLE-4 :Raw Data Classification results of different architectures on Subtask:A test data

| Models | Set | Accuracy | precision | Recall | F1 Score |
|---|---|---|---|---|---|
| Autoencoder | English | 0.78 | 0.68 | 0.78 | 0.70 |
| | Bangla | 0.55 | 0.47 | 0.55 | 0.41 |
| | Hindi | 0.58 | 0.52 | 0.58 | 0.47 |
| LSTM | English | 0.78 | 0.62 | 0.78 | 0.69 |
| | Bangla | 0.55 | 0.30 | 0.55 | 0.39 |
| | Hindi | 0.58 | 0.34 | 0.58 | 0.43 |
| BiLSTM | English | 0.68 | 0.63 | 0.68 | 0.65 |
| | Bangla | 0.57 | 0.45 | 0.57 | 0.50 |
| | Hindi | 0.56 | 0.45 | 0.56 | 0.50 |
| Word2vec | English | 0.78 | 0.68 | 0.78 | 0.72 |
| | Bangla | 0.61 | 0.59 | 0.61 | 0.55 |
| | Hindi | 0.64 | 0.60 | 0.64 | 0.60 |
| BERT | English | 0.79 | 0.79 | 0.79 | 0.79 |
| BERT Multi-Lingual | Bangla | 0.72 | 0.71 | 0.72 | 0.72 |
| BERT Multi-lingual | Hindi | 0.69 | 0.69 | 0.69 | 0.69 |
| gpt2 | English | 0.80 | 0.76 | 0.80 | 0.77 |
| gpt2 | Bangla | 0.73 | 0.74 | 0.73 | 0.73 |
| gpt2 | Hindi | 0.63 | 0.62 | 0.63 | 0.62 |

TABLE-5: Raw data with Augmented Data Classification results of different architectures on Subtask:A test data

| Models | Set | Accuracy | precision | Recall | F1 Score |
|---|---|---|---|---|---|
| Autoencoder | English | 0.66 | 0.67 | 0.66 | 0.67 |
| | Bangla | 0.55 | 0.48 | 0.55 | 0.45 |
| | Hindi | 0.52 | 0.47 | 0.52 | 0.48 |
| LSTM | English | 0.60 | 0.67 | 0.60 | 0.63 |
| | Bangla | 0.54 | 0.47 | 0.54 | 0.49 |
| | Hindi | 0.49 | 0.47 | 0.49 | 0.47 |
| BiLSTM | English | 0.65 | 0.66 | 0.65 | 0.65 |
| | Bangla | 0.44 | 0.48 | 0.44 | 0.42 |
| | Hindi | 0.46 | 0.45 | 0.46 | 0.44 |
| Word2vec | English | 0.71 | 0.73 | 0.71 | 0.72 |
| | Bangla | 0.59 | 0.56 | 0.59 | 0.53 |
| | Hindi | 0.57 | 0.62 | 0.57 | 0.59 |
| BERT | English | 0.75 | 0.78 | 0.75 | 0.77 |
| BERT MultiLingual | Bangla | 0.71 | 0.70 | 0.71 | 0.70 |
| BERT Multilingual | Hindi | 0.68 | 0.69 | 0.68 | 0.68 |
| gpt2 | English | 0.75 | 0.71 | 0.75 | 0.73 |
| gpt2 | Bangla | 0.66 | 0.64 | 0.66 | 0.64 |
| gpt2 | Hindi | 0.63 | 0.62 | 0.63 | 0.62 |

TABLE-6: Machine translated English Data Classification results of different architectures on Subtask:A test data

| Models | Accuracy | precision | Recall | F1 Score |
|---|---|---|---|---|
| Autoencoder | 0.77 | 0.69 | 0.77 | 0.70 |
| LSTM | 0.74 | 0.64 | 0.74 | 0.69 |
| BiLSTM | 0.69 | 0.66 | 0.69 | 0.67 |
| Word2vec | 0.77 | 0.71 | 0.77 | 0.73 |
| BERT | 0.78 | 0.78 | 0.78 | 0.78 |
| gpt2 | 0.76 | 0.76 | 0.76 | 0.76 |

We have observed that the gpt2 model performed best on English and Bangla raw data, whereas the BERT model performed best on Hindi raw data. Gpt2 gave the highest accuracy of 80 percent on English raw data. However, the Bert model performed best on the augmented and machine-translated datasets. It gave 0.78 percent accuracy on the machine-translated dataset. From the experiment, we have found that the BERT model performed best on noisy datasets, while the gpt2 model performed best on the raw dataset that does not contain any noise. It is aparent that, using machine translated data set is quite risky since the dataset contains noises and requires human intervention, which is costly and time-consuming. This work shows that the BERT model can work well on the noisy dataset. We evaluated the model with unseen raw data and got 78 percent accuracy, which is pretty good for industrial and future investigation purposes.

For the fully noisy dataset, the training-validation accuracy curve is repre- sented in Figure 5. For all models, the training accuracy is higher than the validation accuracy, representing that the model has learned the dataset properly without being overfitted or under-fitted.

Figure 5 illustrates a high-level schematic representation of the model accuracy curve which is derived from BERT model using fully machine translated data .

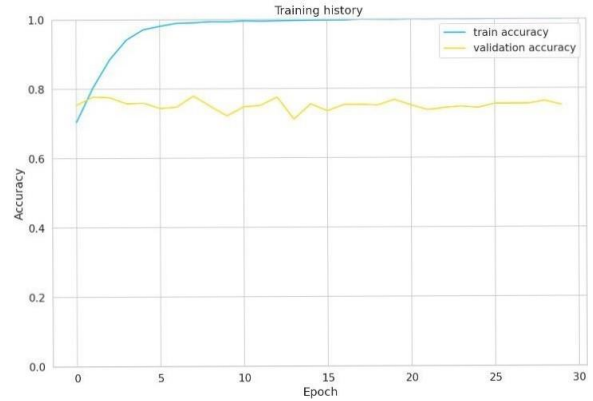

Fig. 5. Curve of model accuracy derived from BERT model using fully machine translated dataset

The confusion matrix shows the number of True positive and False negative results has been predicted by each of the model. Figure 6 illustrates a high-level schematic representation of the Confusion matrix for Bert model using fully machine translated.

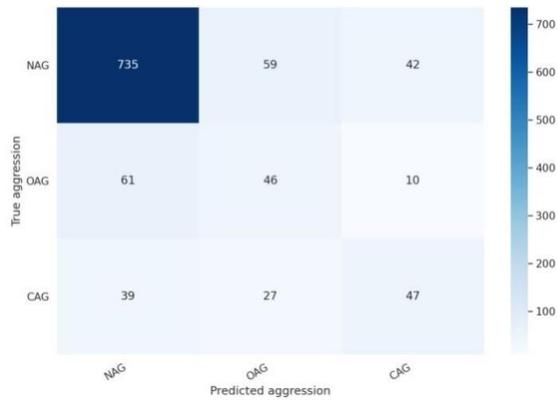

Fig. 6. Confusion matrix for Bert model using fully machine translated dataset.

## V. CONCLUSION

In this work, we have presented the process of generating machine-translated data and how the deep learning models perform on this dataset. We have made a comparative analysis with the performance of the models on raw data and semi-noisy datasets. The raw data denotes the dataset we have collected from the organization. The semi- noisy dataset denotes the augmented data we have added with raw data. We have used models such as LSTM, BiLSTM, LSTM-Autoencoder, word2vec, BERT, and GPT-2 and evaluated the performance of the models using performance metrics such as Accuracy, F1-score, Precision, and recall. The Performance metric shows that the BERT model performed Best on the Machine translated and semi- noisy data, and the gpt2 model performed best on the raw dataset. The difference between the accuracy on raw, semi-noisy, and Machine translated is minimal. The highest accuracy for raw data is 80 percent, for semi-noisy data is 78 percent, and for machine-translated 78 percent. It is clear that training the BERT model using machine-translated data gives almost the same result as the raw dataset, which may be useful for the dataset that lacks a large dataset. Using our approach, future researchers will be able to analyze various problems associated with text datasets that were left behind due to the availability of datasets availability.
## ACKNOWLEDGEMENT

This work is partially supported by the National Science Foundation Award #2100115. Any opinions, findings, and conclusions or recommendations expressed in this material are those of the authors and do not necessarily reflect the views of the National Science

## REFERENCES

[1] C. E. Notar, S. Padgett, and J. Roden, "Cyberbullying: A review of the literature.," *Universal journal of educational research*, vol. 1, no. 1, pp. 1–9, 2013.

[2] B. Auxier and M. Anderson, "Social media use in 2021," *Pew Research Center*, vol. 1, pp. 1–4, 2021.

[3] W. Akram and R. Kumar, "A study on positive and negative effects of social media on society," *International Journal of Computer Sciences and Engineering*, vol. 5, no. 10, pp. 351–354, 2017.

[4] A. M. Memon, S. G. Sharma, S. S. Mohite, and S. Jain, "The role of online social networking on deliberate self-harm and suicidality in adolescents: A systematized review of literature," *Indian journal of psychiatry*, vol. 60, no. 4, p. 384, 2018.

[5] J. L. Trotter and N. E. Allen, "The good, the bad, and the ugly: Domestic violence survivors' experiences with their informal social networks," *American journal of community psychology*, vol. 43, no. 3, pp. 221–231, 2009.

[6] H. Rosa, N. Pereira, R. Ribeiro, P. C. Ferreira, J. P. Carvalho, S. Oliveira, L. Coheur, P. Paulino, A. V. Simão, and I. Trancoso, "Automatic cyberbullying detection: A systematic review," *Computers in Human Behavior*, vol. 93, pp. 333–345, 2019.

[7] B. S. Nandhini and J. Sheeba, "Online social network bullying detection using intelligence techniques," *Procedia Computer Science*, vol. 45, pp. 485–492, 2015.

[8] K. R. R. Turjo, P. A. D'Costa, S. Bhowmick, A. Galib, S. Raian, M. S. Akter, N. Ahmed, and M. Mahdy, "Design of low-cost smart safety vest for the prevention of physical abuse and sexual harassment," in *2021 24th International Conference on Computer and Information Technology (ICCIT)*, pp. 1–6, IEEE, 2021.

[9] A. Ioannou, J. Blackburn, G. Siringhini, E. De Chrisiofaro, N. Kouriellis, M. Sirivianos, and P. Zaphiris, "From risk factors to detection and intervention: A metareview and practical proposal for research on cyberbullying," in *2017 IST-Africa Week Conference (IST-Africa)*, pp. 1–8, IEEE, 2017.

[10] A. Al Mazari, "Cyber-bullying taxonomies: Definition, forms, consequences and mitigation strategies," in *2013 5th International Conference on Computer Science and Information Technology*, pp. 126–133, IEEE, 2013.

[11] A. Perera and P. Fernando, "Accurate cyberbullying detection and prevention on social media," *Procedia Computer Science*, vol. 181, pp. 605–611, 2021.

[12] A. Campan, A. Cuzzocrea, and T. M. Truta, "Fighting fake news spread in online social networks: Actual trends and future research directions," in *2017 IEEE International Conference on Big Data (Big Data)*, pp. 4453–4457, IEEE, 2017.

[13] C. Evangelio, P. Rodriguez-Gonzalez, J. Fernandez-Rio, and S. Gonzalez-Villora, "Cyberbullying in elementary and middle school students: A systematic review," *Computers & Education*, vol. 176, p. 104356, 2022.

[14] S. K. Schneider, L. O'donnell, A. Stueve, and R. W. Coulter, "Cyberbullying, school bullying, and psychological distress: A regional census of high school students," *American journal of public health*, vol. 102, no. 1, pp. 171–177, 2012.

[15] O. Sharif and M. M. Hoque, "Tackling cyber-aggression:

Identification and fine-grained categorization of aggressive texts on social media using weighted ensemble of transformers," *Neurocomputing*, vol. 490, pp. 462–481, 2022.

[16] J. Lv, H. Ren, X. Guo, C. Meng, J. Fei, H. Mei, and S. Mei, "Nomogram predicting bullying victimization in adolescents," *Journal of affective disorders*, vol. 303, pp. 264–272, 2022.

[17] M. Alotaibi, B. Alotaibi, and A. Razaque, "A multichannel deep learning framework for cyberbullying detection on social media," *Electronics*, vol. 10, no. 21, p. 2664, 2021.

[18] M. F. Ahmed, Z. Mahmud, Z. T. Biash, A. A. N. Ryen, A. Hossain, and F. B. Ashraf, "Cyberbullying detection using deep neural network from social media comments in bangla language," *arXiv preprint arXiv:2106.04506*, 2021.

[19] H. Simon, B. Y. Baha, and E. J. Garba, "Trends in machine learning on automatic detection of hate speech on social media platforms: A systematic review," *FUW Trends in Science & Technology Journal*, vol. 7, no. 1, pp. 001–016, 2022.

[20] S. A. Castaño-Pulgarín, N. Suárez-Betancur, L. M. T. Vega, and H. M. H. López, "Internet, social media and online hate speech. systematic review," *Aggression and Violent Behavior*, vol. 58, p. 101608, 2021.

[21] S. G. Roy, U. Narayan, T. Raha, Z. Abid, and V. Varma, "Leveraging multilingual transformers for hate speech detection," *arXiv preprint arXiv:2101.03207*, 2021.

[22] S. Sadiq, A. Mehmood, S. Ullah, M. Ahmad, G. S. Choi, and B.-W. On, "Aggression detection through deep neural model on twitter," *Future Generation Computer Systems*, vol. 114, pp. 120–129, 2021.

[23] A. Kumar and N. Sachdeva, "A bi-gru with attention and capsnet hybrid model for cyberbullying detection on social media," *World Wide Web*, vol. 25, no. 4, pp. 1537–1550, 2022.

[24] K. S. Alam, S. Bhowmik, and P. R. K. Prosun, "Cyberbullying detection: an ensemble based machine learning approach," in *2021 third international conference on intelligent communication technologies and virtual mobile networks (ICICV)*, pp. 710–715, IEEE, 2021.

[25] A. Desai, S. Kalaskar, O. Kumbhar, and R. Dhumal, "Cyber bullying detection on social media using machine learning," in *ITM Web of Conferences*, vol. 40, p. 03038, EDP Sciences, 2021.

[26] S. Bhattacharya, S. Singh, R. Kumar, A. Bansal, A. Bhagat, Y. Dawer, B. Lahiri, and A. K. Ojha, "Developing a multilingual annotated corpus of misogyny and aggression," *arXiv preprint arXiv:2003.07428*, 2020.

[27] D. Mandic and J. Chambers, *Recurrent neural networks for prediction: learning algorithms, architectures and stability*. Wiley, 2001.

*[28]* M. S. Akter, H. Shahriar, R. Chowdhury, and M. Mahdy, "Forecasting the risk factor of frontier markets: A novel stacking ensemble of neural network approach," *Future Internet*, vol. 14, no. 9, p. 252, 2022.

[29] S. Hochreiter and J. Schmidhuber, "Long short-term memory," *Neural computation*, vol. 9, no. 8, pp. 1735–1780, 1997.

[30] J. Schmidhuber, "A fixed size storage o (n 3) time complexity learning algorithm for fully recurrent continually running networks," *Neural Computation*, vol. 4, no. 2, pp. 243–248, 1992.

[31] H. Song, J. Dai, L. Luo, G. Sheng, and X. Jiang, "Power transformer operating state prediction method based on an lstm network," *Energies*, vol. 11, no. 4, p. 914, 2018.

[32] T. Fischer and C. Krauss, "Deep learning with long short-term memory networks for financial market predictions," *European Journal of Operational Research*, vol. 270, no. 2, pp. 654–669, 2018.

[33] S. Liu, G. Liao, and Y. Ding, "Stock transaction prediction modeling and analysis based on lstm," in *2018 13th IEEE Conference on Industrial Electronics and Applications (ICIEA)*, pp. 2787–2790, IEEE, 2018.

[34] Y. Wang, M. Huang, X. Zhu, and L. Zhao, "Attention-based lstm for aspect-level sentiment classification," in *Proceedings of the 2016 conference on empirical methods in natural language processing*, pp. 606–615, 2016.

[35] J. Du, Y. Cheng, Q. Zhou, J. Zhang, X. Zhang, and G. Li, "Power load forecasting using bilstm-attention," in *IOP Conference Series: Earth and Environmental Science*, vol. 440, p. 032115, IOP Publishing, 2020.

[36] H. Nguyen, K. P. Tran, S. Thomassey, and M. Hamad, "Forecasting and anomaly detection approaches using lstm and lstm autoencoder techniques with the applications in supply chain management," *International Journal of Information Management*, vol. 57, p. 102282, 2021.

[37] K. W. Church, "Word2vec," *Natural Language Engineering*, vol. 23, no. 1, pp. 155–162, 2017.

[38] E. Alsentzer, J. R. Murphy, W. Boag, W.-H. Weng, D. Jin, T. Naumann, and M. McDermott, "Publicly available clinical bert embeddings," *arXiv preprint arXiv:1904.03323*, 2019.

[39] M. Munikar, S. Shakya, and A. Shrestha, "Fine-grained sentiment classification using bert," in *2019 Artificial Intelligence for Transforming Business and Society (AITB)*, vol. 1, pp. 1–5, IEEE, 2019.

[40] J. Devlin, M.-W. Chang, K. Lee, and K. Toutanova, "Bert: Pre-training of deep bidirectional transformers for language understanding," *arXiv preprint arXiv:1810.04805*, 2018.

[41] K. Lagler, M. Schindelegger, J. Böhm, H. Krásná, and T. Nilsson, "Gpt2: Empirical slant delay model for radio space geodetic techniques," *Geophysical research letters*, vol. 40, no. 6, pp. 1069–1073, 2013.

[42] D. S. Depto, S. Rahman, M. M. Hosen, M. S. Akter, T. R. Reme, A. Rahman, H. Zunair, M. S. Rahman, and M. Mahdy, "Automatic segmentation of blood cells from microscopic slides: a comparative analysis," *Tissue and Cell*, vol. 73, p. 101653, 2021.